\documentclass[journal=jacsat,manuscript=article]{achemso}

\usepackage[version=3]{mhchem} 
\usepackage[T1]{fontenc}       

\usepackage{amssymb}
\usepackage{latexsym}

\usepackage{url}
\usepackage{xcolor}
\definecolor{newcolor}{rgb}{.8,.349,.1}

\usepackage{hyperref}

\usepackage[switch,pagewise]{lineno} 
\usepackage{graphicx, subfigure}
\usepackage{tabularx}
\usepackage{multirow}
\usepackage{makecell}
\usepackage{url}
\usepackage{float}
\usepackage{bm}
\usepackage{color}
\usepackage{booktabs}
\usepackage{threeparttable}
\usepackage{xcolor}
\usepackage{algorithm}
\usepackage{algorithmic}

\usepackage{amsmath}

\author{Ling Gao}
\affiliation[School of Information Science and Technology Zhejiang Sci-Tech University]
{School of Information Science and Technology Zhejiang Sci-Tech University, Ningbo, China}

\author{Zhenyu Shu}
\email{shuzhenyu@nit.zju.edu.cn}
\affiliation[School of Computer and Data Engineering NingboTech University]
{School of Computer and Data Engineering NingboTech University,  Ningbo, China}

\author{Shiqing Xin}
\affiliation[School of Computer Science and Technology ShanDong University]
{School of Computer Science and Technology ShanDong University, Qingdao, China}

\title{InfoGNN: End-to-end deep learning on mesh via graph neural networks}


\begin{document}

%
%
%
%
%

\begin{abstract}
  3D models are widely used in various industries, and mesh data has become an indispensable part of 3D modeling because of its unique advantages.
  Mesh data can provide an intuitive and practical expression of rich 3D information.
  However, its disordered, irregular data structure and complex surface information make it challenging to apply with deep learning models directly.
  Traditional mesh data processing methods often rely on mesh models with many limitations, such as manifold, which restrict their application scopes in reality and do not fully utilize the advantages of mesh models.
  This paper proposes a novel end-to-end framework for addressing the challenges associated with deep learning in mesh models centered around graph neural networks (GNN) and is titled InfoGNN.
  InfoGNN treats the mesh model as a graph, which enables it to handle irregular mesh data efficiently. Moreover, we propose InfoConv and InfoMP modules, which utilize the position information of the points and fully use the static information such as face normals, dihedral angles, and dynamic global feature information to fully use all kinds of data.
  In addition, InfoGNN is an end-to-end framework, and we simplify the network design to make it more efficient, paving the way for efficient deep learning of complex 3D models.
  We conducted experiments on several publicly available datasets, and the results show that InfoGNN achieves excellent performance in mesh classification and segmentation tasks.
  
\end{abstract}

\section{Introduction}
Mesh data is widely used to represent three-dimensional models in the field of computer graphics and computer vision. Being highly compatible and versatile,
it is supported by most 3D graphics software and platforms. It can represent various shapes and topologies, from simple objects like cubes and spheres to complex scenes with millions of polygons.
It can also store additional information about the object it represents, such as vertex colors, texture coordinates, and normals, to create more realistic and visually appealing graphics.
Some recent applications of mesh processing and analysis include in the field of
game~\cite{Yoon_Kim_2012},
medicine~\cite{Błaszczyk_Jabbar_Szmyd_Radek_2021},
car shape reconstruction~\cite{Nozawa_Shum_Ho_Morishima_2019}, and
3D content creation~\cite{lin2023magic3d}.
These modern applications require high-level processing of 3D data and demand the use of learning-based approaches that utilize data information and labeled or unlabeled datasets. For instance, self-driving robots make use of 3D meshes to navigate complex environments.
Extracting meaningful features from 3D data is essential to enhance tasks such as object classification segmentation.
Learning-based approaches can accomplish these tasks with state-of-the-art results without the need for complex algorithms and data processing.

This paper primarily focuses on mesh classification and segmentation, two fundamental tasks in mesh processing that are essential for many downstream applications.
Mesh classification involves assigning a category label to a mesh, while mesh segmentation partitions a mesh into semantically meaningful regions.
However, since the 3D data is not regular, applying CNN or other methods to learn the features is difficult.
Early techniques for analyzing 3D shapes use hand-crafted feature descriptors like Gaussian Curvature (GC)~\cite{gal2006salient} and
Average Geodesic Distance (AGD)~\cite{shapira2010contextual}.
These methods are theoretically sound but more computationally expensive, especially for more complex shapes.
With the advent of deep learning, researchers combine manual feature descriptors with deep learning techniques, resulting in more effective 3D shape classification.
For example, Guo et al.~\cite{guo20153d} proposes a method that combines artificial features with traditional CNNs to learn more effective features for 3D shape classification.
However, these approaches are limited because manual features are restricted to manifold shapes.
Some methods involve analyzing 3D shapes by projecting them into 2D using established 2D analysis techniques like Sarkar et al.~\cite{Sarkar_Hampiholi_Varanasi_Stricker_2018}.
While this approach can achieve better results in a mature 2D network, it is limited by the masking problem during the 3D to 2D projection and the complexity of learning projections on intricate models.
In recent years, researchers have proposed innovative methods combining knowledge of computer graphics with deep learning, such as MeshCNN~\cite{hanocka_Meshcnn_2019}, which proposes specialized convolution and pooling layers that operate on the mesh edges make it effective in handling input meshes of different sizes. However, such methods often require specific model information like mesh edges, making them less flexible.
The methods have advantages and disadvantages but mostly require 3D models that meet specific conditions.


Our paper aims to design an end-to-end network to handle 3D models without various constraints and utilize the information of the 3D models to learn better features.
To this end, we design InfoGNN, which is based on DGCNN~\cite{DGCNN} and incorporates additional attribute information of mesh to aggregate neighborhood information more effectively. Furthermore, it can enrich the details by adding global information in advance. 



Our main contributions are summarized as follows:

\begin{itemize}
	\item We proposes a novel framework that enables end-to-end flexible training on mesh without various constraints.
	\item We design the novel InfoConv and InfoMP modules to take full advantage of mesh information and global information.
\end{itemize}

The rest of the paper is organized as follows.
Section~\ref{sec:related} briefly reviews related work on 3D shape analysis.
Section~\ref{sec:method} introduces our proposed framework and the associated InfoConv and InfoMP modules.
Section~\ref{sec:experiments} presents experimental comparison results between our approach and state-of-the-art methods.
Section~\ref{sec:ablation}, we show the results of the comparison experiments and analyze the impact of various factors on the experiments.
Section~\ref{sec:performance} describes the tools and runtime environment used for the implementation.
Finally, Section~\ref{sec:conclusion} concludes this paper by discussing our future work.





\section{Related Work}
\label{sec:related}



\subsection{Hand-crafted feature based Methods}

In early research, feature descriptors are extensively used to capture geometric information about 3D shapes and perform various tasks such as shape matching, retrieval, and segmentation. A variety of artificial feature descriptors are proposed, such as Gaussian Curvature (GC)~\cite{gal2006salient} for surface part matching method in triangular mesh representation, Average Geodesic Distance (AGD)~\cite{shapira2010contextual} for capturing shape context, Shape Diameter Function (SDF)~\cite{Shapira_Shamir_Cohen-Or_2008} for encoding global shape information, Scale-Invariant Heat Kernel Signatures (SIHKS)~\cite{Bronstein_Kokkinos_2010} for capturing shape and topology information. Heat Kernel Signatures (HKS)~\cite{Raviv_Bronstein_Bronstein_Kimmel_2010} for describing local shape properties. However, these artificial feature descriptors are often handcrafted and need more ability to capture complex geometric features.
Researchers combine traditional feature descriptors with deep learning techniques to address these limitations. For example, Guo et al.~\cite{guo20153d} proposes a method that combines artificial features with traditional CNNs to learn more effective features for 3D shape classification. Xie et al.~\cite{Xie_Xu_Liu_Xiong_2014} proposes a method that combines artificial features with Extreme Learning Machines (ELMs) to improve the performance of 3D shape retrieval.
As 3D models become more and more complex, individual geometric features cannot provide enough feature information to segment the model accurately into meaningful parts. In addition, there are significant differences between different categories of 3D models, which cannot be segmented by a single geometric feature alone.
Although these methods can achieve good accuracy, the generation of some feature descriptors requires special conditions, such as 2D flow patterns, and generally takes extra time.
As the volume and complexity of data continue to grow within the extensive data landscape, the limitations of traditional methods render them increasingly unsuitable for future applications. To address this challenge, robust and adaptable techniques are paramount for effectively handling the intricacies of 3D models while maintaining computational efficiency.


\subsection{2D Projection based Methods}

Accomplishes various complex tasks by seamlessly transitioning between the two-dimensional (2D) and three-dimensional (3D) realms. It first projects the intricate 3D shape onto a 2D plane, where well-established 2D techniques can be effectively applied.
Based on years of research and development, these techniques offer proven solutions to a wide range of problems. Once the task is completed in the 2D domain, the results are seamlessly mapped back onto the original 3D shape, preserving the integrity and accuracy of the model.
This unique approach enables the application of powerful 2D methods to complex 3D problems, expanding the capabilities of existing tools and techniques.
For example, Su et al.~\cite{Su_Maji_Kalogerakis_Learned-Miller_2015} are the first to apply a multi-view CNN for the shape classification task; however, this approach cannot perform semantic segmentation.
Evangelos et al.~\cite{Kalogerakis_Averkiou_Maji_Chaudhuri_2016} suggests a more comprehensive multi-view framework for shape segmentation: generating image level segmentation maps per view and then solving for label consistency using CRF.
Sarkar et al.~\cite{Sarkar_Hampiholi_Varanasi_Stricker_2018} represents the 3D shape using multi-layered height maps along with a novel multi-view merging technique.
The good thing about this approach is that ready-made and mature network structures can be used, but masking is a problem.

\subsection{Graph-based Methods}
Such a method is often applied in the point cloud; we can compute the graphical relations between points by finding k-nearest neighbors on the points in the spatial domain.
Shen et al. ~\cite{KCnet} introduce a novel concept of a point-set kernel, which is similar to a convolution kernel for images but specifically designed for 3D point clouds. This kernel enables the network to capture local 3D geometric structures, which is crucial for semantic learning. It proposes a recursive feature aggregation method based on a nearest-neighbor graph constructed from 3D positions. It effectively captures local high-dimensional feature structures, enhancing the network's ability to learn semantic features.
Wang et al.~\cite{DGCNN} introduce a novel neural network module called EdgeConv, which is specifically designed for high-level tasks on point clouds, such as classification and segmentation.
Unlike existing modules that operate in extrinsic space or treat each point independently, EdgeConv dynamically computes graphs in each layer of the network, which allows the model to adapt to the specific structure of the point cloud and capture more complex relationships between points.





\subsection{Other Methods}
MeshCNN~\cite{hanocka_Meshcnn_2019} specialize convolution and pooling operations on edge, which are implemented to simplify the model while preserving the original information and leveraging the unique properties of meshes to process 3D shapes directly. Enable task-driven feature extraction and redundancy reduction.
MeshWalker~\cite{lahav_Meshwalker_2020} introduces a novel approach for representing 3D shapes using random walks on triangular meshes, which is highly efficient, requiring only a small number of training examples.
Li et al.~\cite{TPNet} propose a novel 3D mesh learning method called TPNet, which enhances the perception and learns long-range information by customizing dilated convolution for non-uniform mesh data.
Specifically, a faithful strategy is devised to locate the positions where dilate convolutions accurately can be applied despite the non-uniformity and irregularity of the mesh data.
These methods are entirely based on the various properties of the mesh combined with deep learning to design and train the experiment.

\section{Method}
\label{sec:method}

\begin{figure*}[ht] 
	\centering 
	\includegraphics[width=\textwidth]{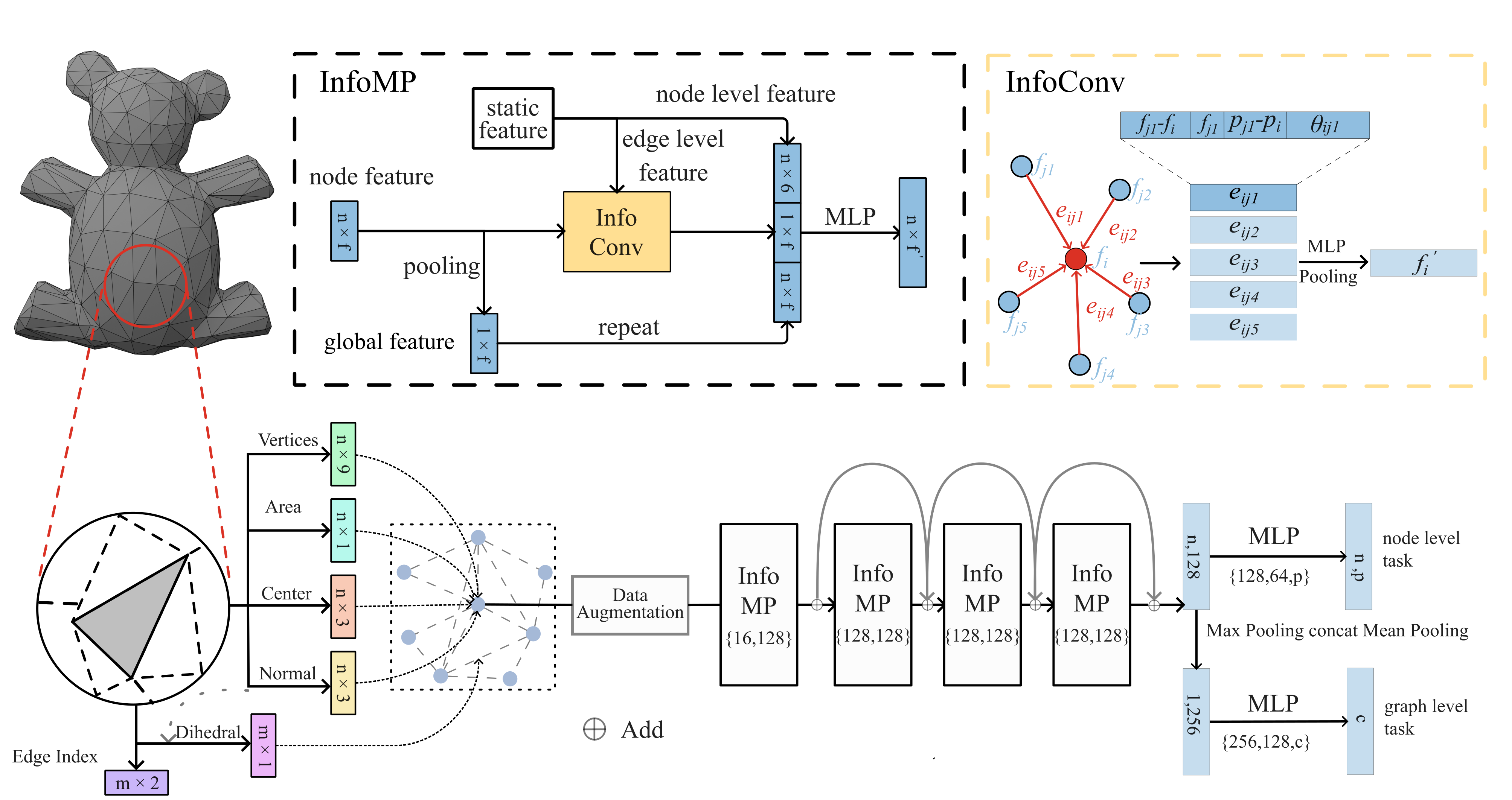} 
	\caption{The input is a 3D mesh model. We preprocess the model, convert it into a unified representation and storage in the form of a graph, and extract and store simple and useful information. After that, we feed this graph data into the message-passing graph neural network InfoMP for feature aggregation and learning to output useful features. The numbers in parentheses represent the input and output dimensions. The hidden layer of InfoMP consists of multiple parts, which are generally twice as large as the input dimension. After 4 InfoMP layers with residual connections, the feature vectors of all nodes are obtained. Depending on the task, a suitable classification head can be chosen. Here, we use a simple multilayer perceptron (MLP) for classification. The numbers in parentheses below represent the input dimensions, hidden layers, and output layers of the MLP.}
	\label{fig:1} 
\end{figure*}

\begin{figure}[htb] 
	\centering 
	\includegraphics[width=0.8\columnwidth]{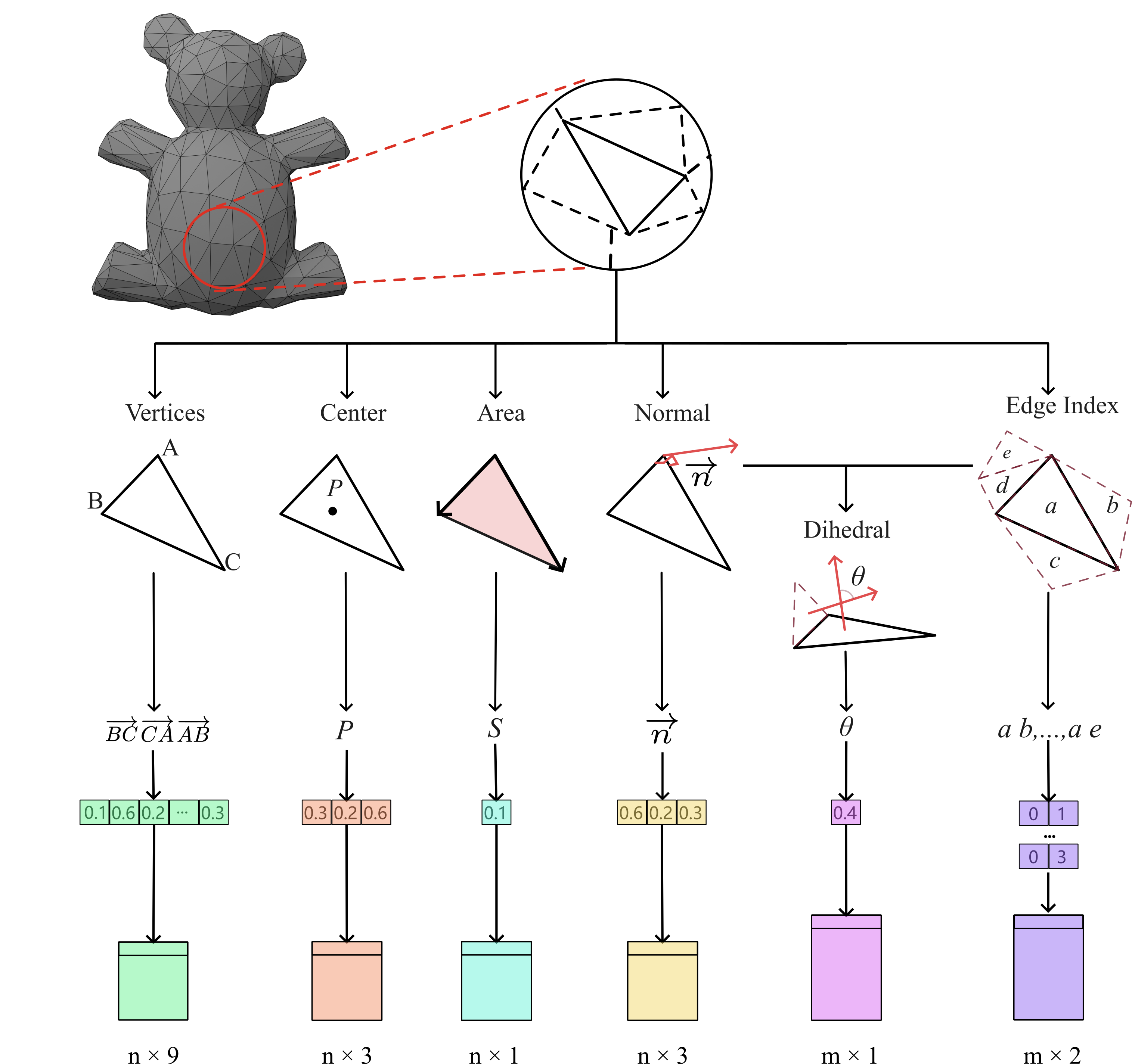} 
	\caption{Model information data composition. The graph data of each model contains six different types of information that can be categorized into node information and edge information.
	The node information includes the original point coordinate information inherent in each model. Information such as face center coordinates, face normal vector, and area can be calculated based on the original point coordinates.
	The edge information is constructed from the face center coordinates using the KNN algorithm, which has the advantages of time-saving and adjustability.
	Combining the face normal vector and edge information, the simple feature information of the edge, such as the dihedral angle, can be further calculated.}

	\label{fig:2} 
\end{figure}

In this section, we present the design of InfoGNN.
The overall architecture also illustrated in Figure~\ref{fig:1}.
We are inspired by PointNet~\cite{charles_pointnet_2017} and DGCNN~\cite{DGCNN}, which tries to learn 3D features end-to-end.
However, it differs from PointNet, which only focuses on individual points. We construct edges of neighboring nodes using the KNN algorithm and compute the neighborhood information using a graph neural network. We also add cross-domain connections to enrich the types of edges.
Unlike DGCNN, which reconstructs edges at each layer, we only construct edges during preprocessing.
Compared with using KNN at each layer, our method is more time and power-efficient.
Compared to the point cloud, the mesh has point and surface data, and we want to use these data better. We calculate valuable attributes, such as area and dihedral angle, and incorporate them into the network.
The init data composition is shown in Figure~\ref{fig:2}.
During training, we add the global features from the previous layer to the current layer to improve the performance.



\subsection{Data Process}

Common mesh data  is mainly composed of vertex set and face set, we are based on the triangular mesh composition of the 3D shape of the data, the data can be expressed as follows:
\begin{equation}
	P = \{p_i  | i=1,2,\cdots ,N  \} \in \mathbb{R}^{N \times 3},
\end{equation}
\begin{equation}
	F = \{f_i | i=1,2,\cdots , M\} \in \mathbb{N}^{M \times 3},
\end{equation}
where $p_i$ is the position information of the $i$-th point, usually expressed as three floating-point numbers.
$N$ indicates the number of points in a 3D model.
$M$ indicates the number of faces in a 3D model.
$P$ then denotes the set of points in a 3D model.
$f_i$ is the face information of the 3D model, which contains the indexes of three $p_i$ and specifies that a triangular face is constructed counterclockwise, which implies the direction of the face.
Because the labels of the mesh dataset generally correspond to the faces, in order to unify the operation, we abstract the faces into nodes for training.

So, we first calculate the center position of the face based on the information about the face information and the relevant point.
We write this as follows:
\begin{equation}
	P^m = \{p^m_i  | i=1,2,\cdots ,M  \} \in \mathbb{R}^{M \times 3},
\end{equation}
where $p^m_i$ is the center position of $i$-th face and is generally represented by three floating-point numbers like point position.

$P^m$ then denotes the set of faces in a 3D model.
Due to the large amount of data and irregularity in 3D models, some experiments are performed by sampling fixed points.
Our experiments are also done by sampling according to the needs.
In getting $p^m_i$, we also calculate the area, face normal, and dihedral angle to enrich the input feature. We call them the static feature and mark them as $S,n,\theta $ respectively.

Since our framework does not impose data constraints on the mesh model, we are able to apply a richer set of data enhancements, including not only traditional rotation and scaling, but also cropping and perturbation operations.
This  can improves the robustness and generalization of the model.

As for the edge information, although we can use face information to construct the edge information through methods such as dual graph algorithm, which allows better construction of edges, these methods generally have high time complexity and have no GPU accelerated implementations.
So, we choose to use the KNN algorithm, commonly used in point cloud processing, to construct the edge information.
In addition, we also use the FPS algorithm to sample nodes and construct additional jump links, with the intention of speeding up the propagation of information and enriching the range of local connections so that edges connect not only to surrounding points but also to some more distant points.

The KNN algorithm will efficiently find $k$ neighboring nodes according to $p^m_i$, which gives us the edge information.
As a result, we convert each face to a node in the graph and define the edge as the dependency of information between two faces.
Generally, a directed graph with nodes $V$ and edges $E$ is constructed as follows:
\begin{equation}
	G = (V,E),  V=\{1,2,\cdots,M\} ,E \in V \times V ,
\end{equation}
where $V$ represents the set of all nodes in the graph.
It can also be said to be the set of all face indexes in a 3D model.
$E$ is the set of neighboring node pairs, representing the edge information of the graph,
Suppose $x_i$ is a node in the graph, then
$N(i) = {j:(i,j) \in E }$ is the set of neighboring nodes of $x_i$.
We define the node feature in layer $l$ as:
\begin{equation}
	F^m_l = \{f^m_{li} | i=1,2,\cdots , M\} \in \mathbb{R}^{M \times D}, l \in \{1,2,3,4,5\} ,
\end{equation}
here, $f^m_{li}$ represent the feature of $i$-th node in layer $l$ and have $D$-dimension.
In particular, the first layer is usually a combination of information that serves as the initial input feature.
Which can be demonstrated as follows:

\begin{equation}
	f^m_{l1}  = p^m_i \oplus S_i \oplus n_i ,
\end{equation}
where $\oplus$ represents the concatenate operation so that the node-level information is concated together as input to the neural network.

\subsection{InfoConv}

As shown in Figure~\ref{fig:1}, our InfoConv module is at top right.
The InfoConv module is similar to EdgeConv in that it aggregates the edge features associated with a point and updates the local information to learn dynamic local features.
The difference is that we also include the information of dihedral angle, which is not in the original method, this can enrich the information source and improve the accuracy of the model on one hand, and improve the learning efficiency of the model on the other hand.

Since we are constructing a graph, we can define the edge features as follows:
\begin{equation}
	e_{ij} = h_{ \Theta}(f^m_i,f^m_j,p^m_i,p^m_j,\theta_{ij}) ,
\end{equation}
where $f^m_i$,$f^m_j$ is  the  feature of $i$-th and  $i$-th  node.
$\theta_{ij}$ represents corresponding dihedral angle.
$h_{\Theta} : \mathbb{R}^{D'} \times \mathbb{R}^{D'}$ is a nonlinear function with a learnable parameter $\Theta$.
It can be expressed as follows:

\begin{small}
	\begin{equation}
		\begin{split}
		h_{\Theta}(f^m_i,f^m_j,p^m_i,p^m_j,\theta_{ij}) =& \\ 
		MLP( (1+ e^{-\frac{\left\| p^m_j -p^m_i \right\|_2 }{3}}  + \theta_{ij} &)((f^m_j-f^m_j)  \oplus f^m_j \oplus (p^m_j -p^m_i) \oplus \theta_{ij} )  ) ,
		\end{split}
	\end{equation}
\end{small}\noindent
where $MLP$ stands for multilayer perceptual layer in neural networks, and in this paper, methods such as BatchNorm, LeakyRelu, etc., are also generally implied.

After that, we apply the symmetric function $\mathbb{F} $(usually $max$) to all the edge features corresponding to each node. Then, we can get the dynamic local features of each node $F^m_{lni}$ in $l$-th layer as follows:
\begin{equation}
	F^m_{lni} =\mathbb{F}   (e_{ij}),j\in N(i)
\end{equation}

\subsection{InfoMP}

Figure~\ref{fig:1} on the top left shows our InfoMP module based on a message passing graph neural network.
Compared to the original EdgeConv method, we would like to add different information to the network learning to improve the network's performance.
In summary, we can classify information into four types of features: static features, current features, dynamic global features, and dynamic local features.
Static features are the original input features used to train the neural network model.
Current features are the input features for each InfoMP layer.
Dynamic global features are individual features aggregated into a mesh model's current features using pooling methods.
Finally, dynamic local features are the output of InfoConv, which aggregates information around each point based on edge information.

To be more specific, static features are the original input feature data such as point position, area, and other point-level features. These features can be added as absolute positional parameters to improve the network's ability to recognize target information and enhance its robustness.
It can be expressed as follows:
\begin{equation}
	F^m_{si} = p^m_i \oplus n_i \oplus S_i  , i=1,2,\cdots , M ,
\end{equation}

As for the dynamic global features, considering the performance issue, we use a simpler method, i.e., we compute the dynamic global features of the $l$-th layer by applying a symmetric function $\mathbb{F} $ (usually max) to all the node features in the graph:

\begin{equation}
	F^m_{lg} =\mathbb{F}  (f^m_{li}\times (1+S_i)) , i=1,2,\cdots , M ,
\end{equation}

We add $S$ as a bias in the hope that the neural network will also take the size of the surface into account.
Note that the dynamic global feature here is relative to a single model, i.e., one model corresponds to one dynamic global feature.
For subsequent training, we also need to align the nodes at the data level by repetition.

Finally, due to the different distribution of various information, direct splicing is the best choice, so updating the next layer of features can be expressed as follows:
\begin{equation}
	F^m_{l+1}{i} = MLP( F^m_{lg} \oplus F^m_{lni} \oplus F^m_{si} )
\end{equation}





\section{Experiments}
\label{sec:experiments}
\subsection{Part Segmentation}

\subsubsection{Human Body Segmentation}



The dataset for this study comprises 370 training models
derived from three distinct sources:
SCAPE~\cite{SCAPE}, FAUST~\cite{FAUST}, and MIT~\cite{Vlasic2008ArticulatedMA}.
This collection of models represents a diverse range of human body shapes and forms, providing a comprehensive basis for training and evaluating segmentation algorithms. 18 human meshes from SHREC 07~\cite{SHREC07} are employed for the testing phase.
According to the methodology outlined by Kalogerakis et al.~\cite{Kalogerakis2010}, these meshes are manually segmented into eight distinct labeled segments. This segmentation scheme ensures consistency in labeling and facilitates a more accurate comparison of segmentation results across different algorithms.
To evaluate the effectiveness of our segmentation algorithm, we adopted the measure of correct face classification as proposed in
SNGC~\cite{SNGC}.
This metric assesses the algorithm's ability to accurately identify and label facial regions within the mesh. By employing this well-established evaluation criterion, we can ensure that our results are comparable to those of previous studies and provide a meaningful assessment of our algorithm's performance.

\begin{figure*}[t] 
	\centering 
	\includegraphics[width=0.9\textwidth]{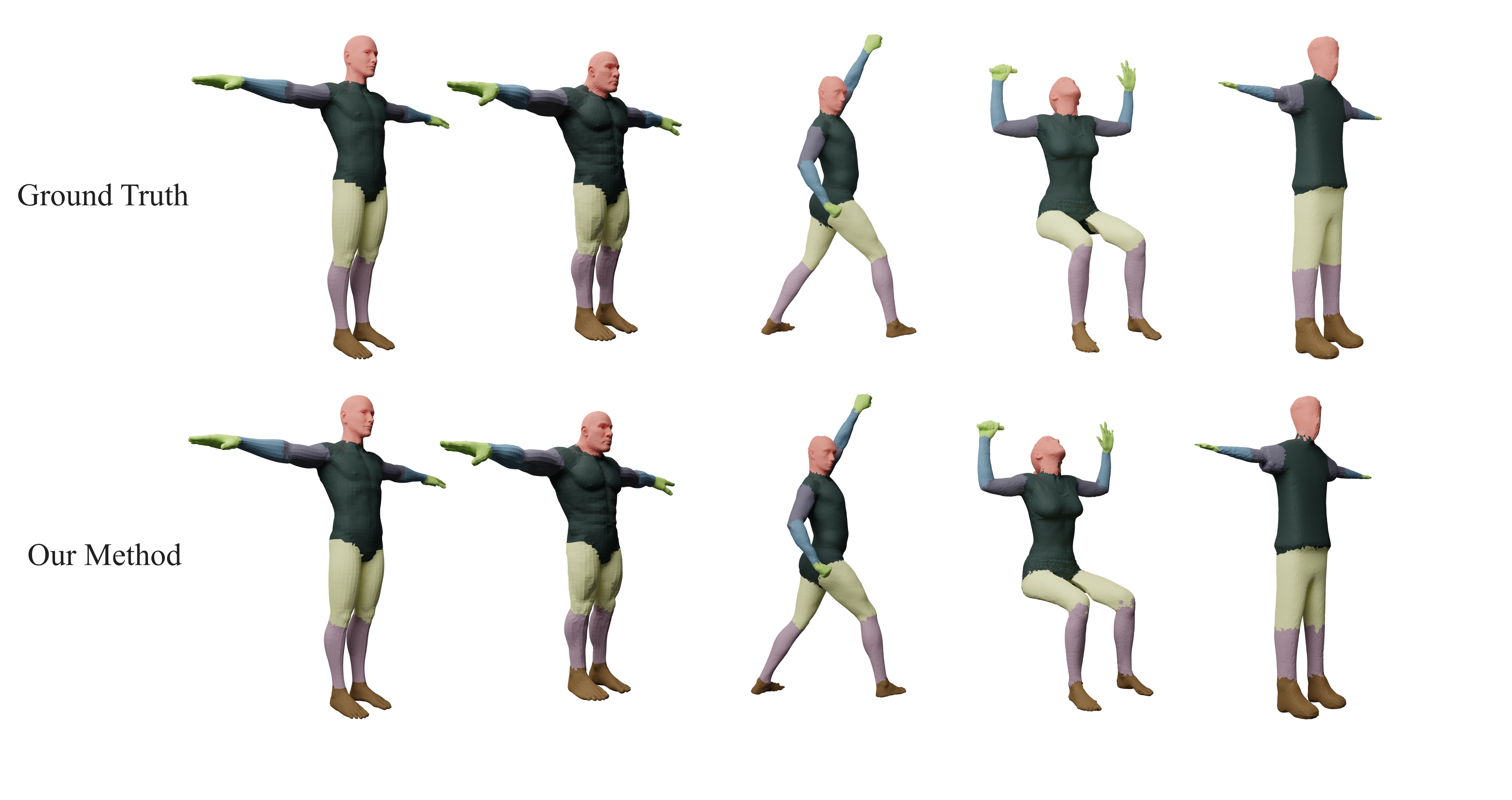}
	\caption{Some samples from the human body dataset and their corresponding renderings of the segmentation results. As can be seen, our segmentation results are mostly accurate, but there are still some boundary divergences that we aim to improve in the future.} 
	\label{fig:3}
\end{figure*}

In order to provide a comprehensive overview of the segmentation results, we compile the findings into Table~\ref{table:hb1}. This table presents the correct face classification rates achieved by our algorithm alongside those obtained by previous works. A comparison of these results reveals that our method surpasses the performance of existing algorithms, demonstrating its state-of-the-art capabilities in human body segmentation.
As shown in Figure~\ref{fig:3}, we visualize the dataset and our results.


\begin{figure*}[ht] 
	\centering 
	\includegraphics[width=0.9\textwidth]{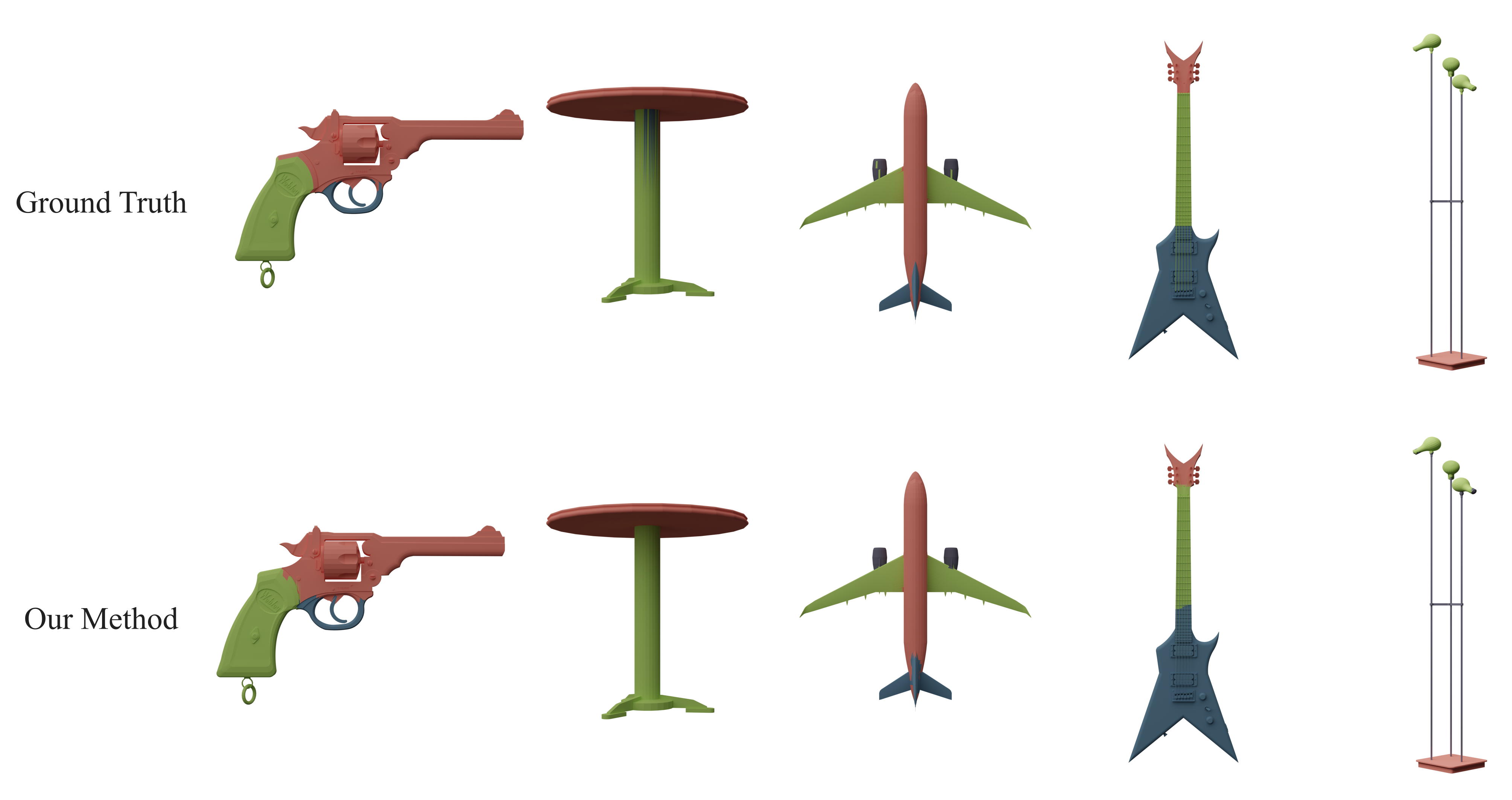} 
	\caption{Some samples from ShapeNetCore and their corresponding renderings of the segmentation results.It can be seen that this dataset segmentation quality is not high and there are some incorrect annotations, but our model still gives results that are more in line with human intuition. This suggests that our model learns better features and is able to effectively utilize the other correct annotations in the dataset for more rational classification.} 
	\label{fig:4} 
\end{figure*}

\begin{table}[t]\footnotesize
	\centering

	\caption{Human Body segmentation result.}
	\label{table:hb1}
	\begin{tabular}{lcc}
		\toprule

		Method                                  & Input       & Face Accuracy(\%) \\

		\midrule
		Our Method                              & Mesh        & \textbf{93.2}     \\
		MeshWalker~\cite{lahav_Meshwalker_2020} & Mesh        & 92.7              \\
		MeshCNN~\cite{hanocka_Meshcnn_2019}     & Mesh        & 89.0              \\
		SNGC~\cite{SNGC}                       & Mesh        & 91.02             \\
		Toric Cover~\cite{ToricCover}           & Mesh        & 88.00             \\
		GCNN~\cite{GCNN}                       & Mesh        & 86.40             \\
		MDGCNN~\cite{MDGCNN}                    & Mesh        & 89.47             \\
		DGCNN~\cite{DGCNN}                      & Point cloud & 89.72             \\
		PointNet++~\cite{PointNetpp}            & Point cloud & 90.77             \\
		\bottomrule
	\end{tabular}
\end{table}

\begin{table}[t]\footnotesize
	\centering
    
	\caption{ModelNet40 classification result.}
	\label{table:mn}
	\begin{tabular}{lcc}
		\toprule

		Method                                 & Input       & Accuracy(\%)   \\
		\midrule
		Our Mehod                              & Mesh        & \textbf{92.02} \\
		MeshNet~\cite{MeshNet}                & Mesh        & 91.9           \\
		SNGC~\cite{SNGC}                      & Mesh        & 91.6           \\
		PointNet~\cite{charles_pointnet_2017} & Point cloud & 89.2           \\
		CrossMoCo~\cite{10229841}              & Point cloud & 91.49          \\

		\bottomrule
	\end{tabular}
\end{table}

\subsubsection{Details}





We use the Adam optimizer with an initial learning rate of 3e-3, epsilon of 1e-6, and weight decay of 1e-4.
Additionally, we dynamically adjust the learning rate based on the mean training epoch loss using a ReduceOnPlateau scheduler with a factor of 0.9, the patience of 3 epochs, and a minimum learning rate of 1e-6, threshold\_mode rel and threshold 1e-3, the scheduler reduces the learning rate by 10\% whenever the mean training epoch loss is not lower than the previous minimum value for three consecutive epochs.
Typically, we employ a batch size of 24 to balance computational efficiency with training effectiveness.
In order to combat overfitting, a common challenge in machine learning, we implement data augmentation techniques such as random scaling, rotation, and cropping of the 3D model, introducing variations in the training data that enhance the model's generalization ability.

\subsubsection{ShapeNetCore Segmentation}

\begin{table*}[t]\footnotesize
	\caption{ShapeNetCore segmentation result.}
	\label{table:snc}
 	\resizebox{\linewidth}{!}{

		\begin{tabular}[width=0.85\columnwidth]{llllllllllllllllll}
			\toprule

			~                                      & mIoU(\%)      & air. & bag           & cap           & car  & cha. & ear.          & gui. & kni.          & lam.          & lap.          & mot. & mug           & pis.          & roc.          & ska.          & tab.          \\
			shapes                                 &               & 2690 & 76            & 55            & 898  & 3758 & 69            & 787  & 392           & 1547          & 451           & 202  & 184           & 283           & 66            & 152           & 5271          \\
			\midrule
			PointNet~\cite{charles_pointnet_2017} & 83.7          & 83.4 & 78.7          & 82.5          & 74.9 & 89.6 & 73.0          & \textbf{91.5} & 85.9          & 80.8          & 95.3          & 65.2 & 93.0          & 81.2          & 57.9          & 72.8          & 80.6          \\
			PointNet++~\cite{PointNetpp}          & 85.1          & 82.4 & 79.0          & 87.7          & 77.3 & 90.8 & 71.8          & 91.0 & 85.9          & 83.7          & 95.3          & 71.6 & 94.1          & 81.3          & 58.7          & 76.4          & 82.6          \\
			Kd-Net~\cite{KdNet}                   & 82.3          & 80.1 & 74.6          & 74.3          & 70.3 & 88.6 & 73.5          & 90.2 & 87.2          & 81.0          & 94.9          & 57.4 & 86.7          & 78.1          & 51.8          & 69.9          & 80.3          \\

			PCNN~\cite{PCNN}                      & 85.1          & 82.4 & 80.1          & 85.5          & 79.5 & 90.8 & 73.2          & 91.3 & 86.0          & 85.0          & \textbf{95.7} & 73.2 & 94.8          & 83.3          & 51.0          & 75.0          & 81.8          \\
			3D-GCN~\cite{Lin_2020_CVPR}            & 85.1          & 83.1 & 84.0          & 86.6          & 77.5 & 90.3 & 74.1          & 90.9 & 86.4          & 83.8          & 95.3          & 65.2 & 93.0          & 81.2          & 59.6          & 75.7          & 82.8          \\
			DGCNN~\cite{DGCNN}                    & 85.2          & 84.0 & 83.4          & 86.7          & 77.8 & 90.6 & 74.7 & 91.2 & 87.5 & 82.8          & \textbf{95.7}          & 66.3 & \textbf{94.9} & 81.1          & 63.5 & 74.5          & 82.6          \\

			Pra-net~\cite{cheng_pra-net_2021}      & 86.3 & 84.4 & \textbf{86.8} & 89.5          & 78.4 & \textbf{91.4} & \textbf{76.4}          & \textbf{91.5} & 88.2          & 85.3          & \textbf{95.7}          & \textbf{73.4} & 94.8          & 82.1          & 62.3          & 75.5          & \textbf{84.0} \\

			Ours                                   & \textbf{86.4}          & \textbf{85.1} & 85.2          & \textbf{91.1} & \textbf{78.9} & 91.2 & 75.6          & 91.3 & \textbf{88.3}          & \textbf{87.5} & 95.5          & 72.1 & 94.5          & \textbf{84.6} & \textbf{64.0}          & \textbf{81.9} & 83.0          \\
			\bottomrule
		\end{tabular}
	 }
\end{table*}

We also employ the ShapeNetCore dataset~\cite{shapenet2015}, a meticulously curated subset of the ShapeNet dataset.
This dataset comprises 16,881 3D models spanning 16 distinct categories, each meticulously cleaned and manually annotated with category and alignment information. Among these models, 14,006 are designated for training, while the remaining 2,874 serve as a testing set. The dataset encompasses 50 part tags, with each model encompassing between 2 and 6 parts.
Following the experimental setup outlined in PointNet~\cite{charles_pointnet_2017}, we sample 2,048 points from each shape for analysis.
In this dataset, we evaluate and compare the performance of part segmentation for this dataset with other benchmarks using the mIoU metric.
IoU is calculated for shape by averaging the IoUs of different parts occurring in that shape.
Category IoU is the average IoU of all shapes in that category. Finally, the overall performance (mIoU) is the average IoU of all testing shapes.
The evaluation results are shown in Table~\ref{table:snc}. We also compare them with similar methods, and we can see better results in many places.
We improve the performance in categories with fewer shapes because we can incorporate information and transformations to fill the data gaps.
However, some categories with large amounts of data still need improvement because there is too much data and noise to learn good features.
As shown in Figure~\ref{fig:4}, we visualize the dataset and our results.
It can be seen that even though the ground truth value has some badly labeled parts, the result of our method can predict better-parted labels.

\subsubsection{Details}

We use the Adam optimizer with an initial learning rate of 3e-4, epsilon of 1e-6, and weight decay of 1e-3,
also, we use the ReduceOnPlateau scheduler with a factor of 0.5, patience of 3 epochs, and a minimum learning rate of 1e-6; typically, we employ a batch size of 24 to balance computational efficiency with training effectiveness.
We also use the same data augmentation techniques, such as random scaling, rotation, and cropping of 3D models, to combat the overfitting problem.

\subsection{Mesh Classification}

\subsubsection{ModelNet40}

\begin{figure}[hb] 
	\centering
	\includegraphics[width=0.7\textwidth]{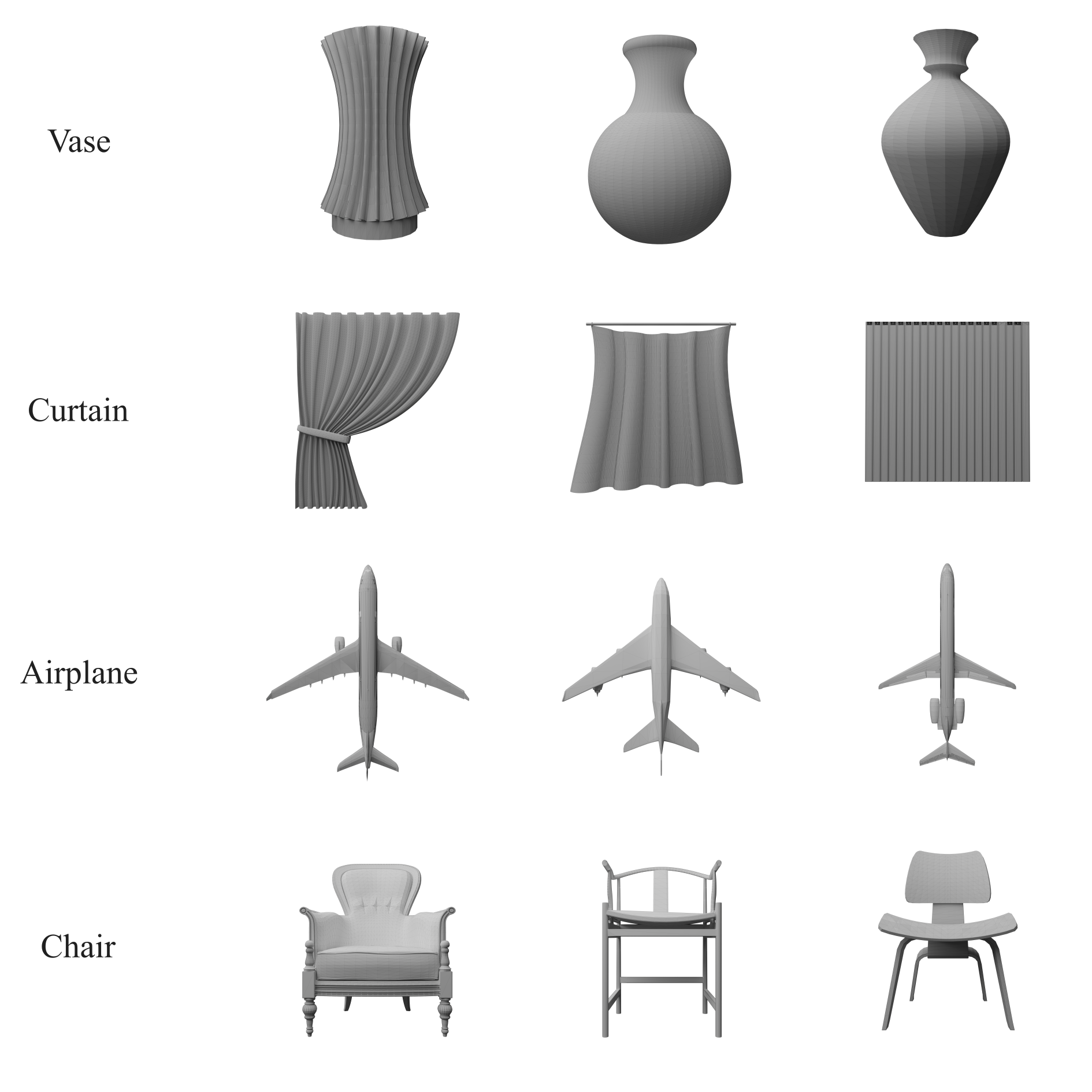 } 
	\caption{Some example shapes from ModelNet40 are used to demonstrate the classification capabilities of our network.} 
	\label{fig:5} 
\end{figure}



For evaluating the classification performance of our model, we utilize the ModelNet40~\cite{modelnet} dataset, a widely recognized benchmark comprising 12,311 CAD models across 40 categories.
Figure~\ref{fig:5} shows some examples of the dataset.
This dataset's extensive collection of models, including 9,843 for training and 2,468 for testing, provides a robust platform for assessing the effectiveness of our approach.
Unlike previous datasets, ModelNet40 presents a more challenging task due to including objects with multiple components and non-watertight meshes, characteristics that pose significant obstacles for specific mesh-based methods.
However, our method's ability to handle models of arbitrary complexity situations is well-suited for this demanding dataset.
Table~\ref{table:mn} shows the comparison of our experimental results with related methods. From the table, we can see that our method achieves better performance than mesh-related methods.
Some of the point cloud based methods  achieve similar results, but their data are usually derived from sampling a 3D mesh model. In order to improve the accuracy of point cloud data using more computationally intensive methods, it is expected to use only a fixed number of points, such as 1024 or 2048.
In contrast, we use the complete mesh format data, which has two problems: the data is noisy, which affects the accuracy of the model, and a large amount of data leads to the limited number of methods available at present. We believe these problems can be solved by optimizing the algorithms and hardware so that the entire mesh format data can be used to more significant potential in relevant tasks.

The excellent performance of multi-view approaches can be found in this dataset, which can be attributed to their reliance on networks pre-trained on a vast repository of images.
However, their effectiveness may not translate to other datasets, such as engraved cubes, and their applicability extends beyond shape analysis tasks such as semantic segmentation.



\subsubsection{Details}

For Modelnet40, we also use the Adam optimizer with an initial learning rate of 1e-3, epsilon of 1e-8, and weight decay of 1e-4, the  ReduceOnPlateau scheduler with the factor of 0.85, the patience of 4 epochs, and a minimum learning rate of 5e-6,
the scheduler reduces the learning rate by 1e-4 whenever the validation loss plateaus for three consecutive epochs.
We employ a batch size of 32 to balance computational efficiency with training effectiveness.
In order to combat overfitting, a common challenge in machine learning, we implement data augmentation techniques such as random scaling, rotation, and cropping of the 3D model, introducing variations in the training data that enhance the model's generalization ability.

\section{Ablation Study}
\label{sec:ablation}

\subsection{Impact of additional information}

\begin{figure}[htb] 
	\centering
	\includegraphics[width=0.8\textwidth]{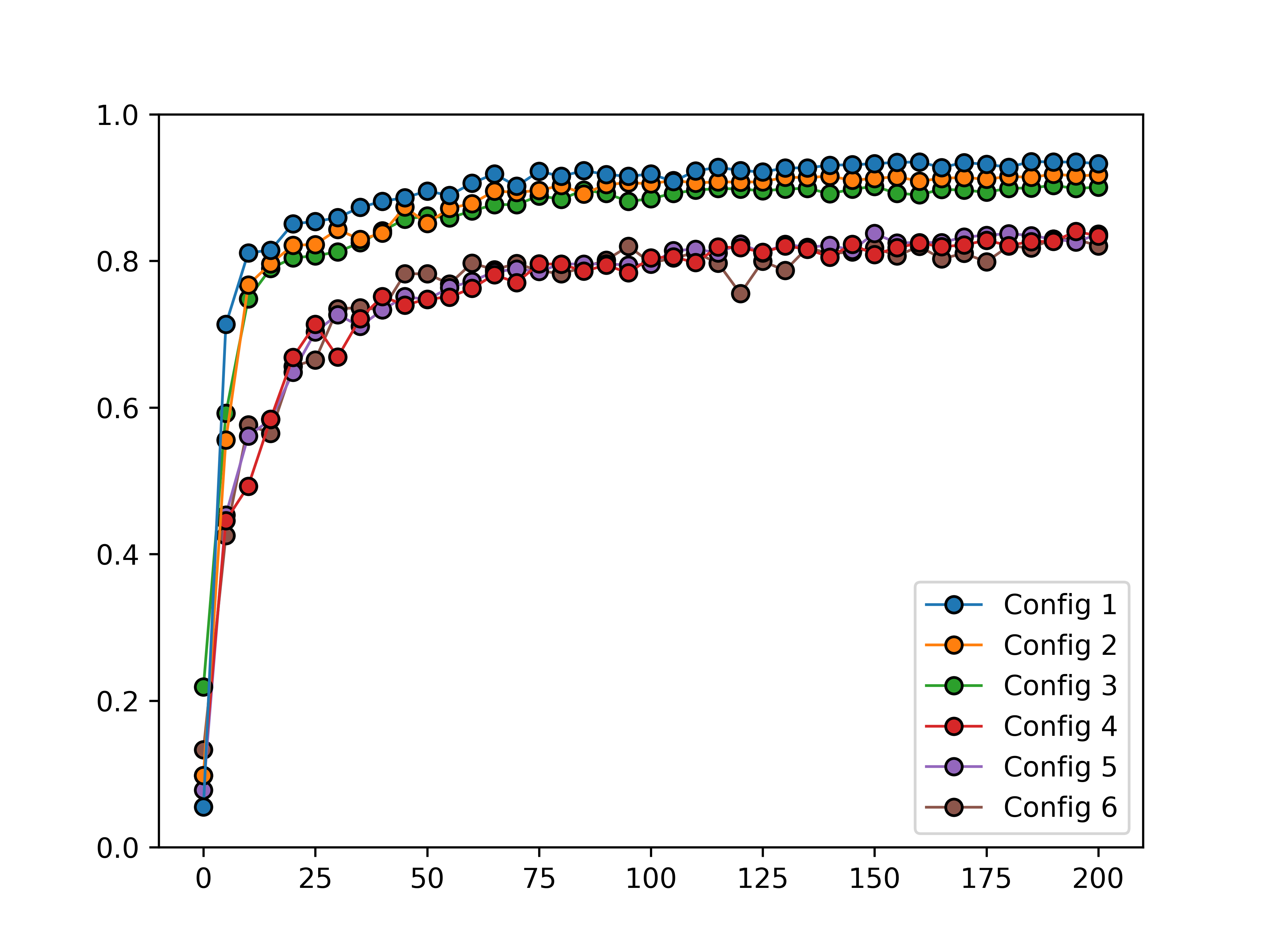} 
	\caption{Experimental results for the test dataset. The horizontal axis represents the number of training rounds, while the vertical axis represents the accuracy of the experiment. Our base configuration (Config 1) includes global features, static information, and all input features. Subsequent configurations build upon the previous one as follows: Config 2 removes global features, Config 3 removes static information, Config 4 reduces input channels to just the center point, Config 5 increases channels to 256, and Config 6 further increases channels to 512.} 
	\label{fig:abl:1} 
\end{figure}

In this section, we briefly use the human body dataset to explore the impact of different choices on our model.
We use a base model with 1024 samples for ease of ablation study.
Our base model configuration is marked as  Config 1, which contains global features, static information, and all input features. Subsequent Configs build on the previous one.
In Config 2, we remove the global features; in Config 3, we further remove the static information; in Config 4, we reduce the input channels to just the center point; in Config 5, we increase the number of channels to 256; and in Config 6, we further increase the number of channels to 512.

\begin{figure}[ht] 
	\centering 
	\includegraphics[width=0.7\textwidth]{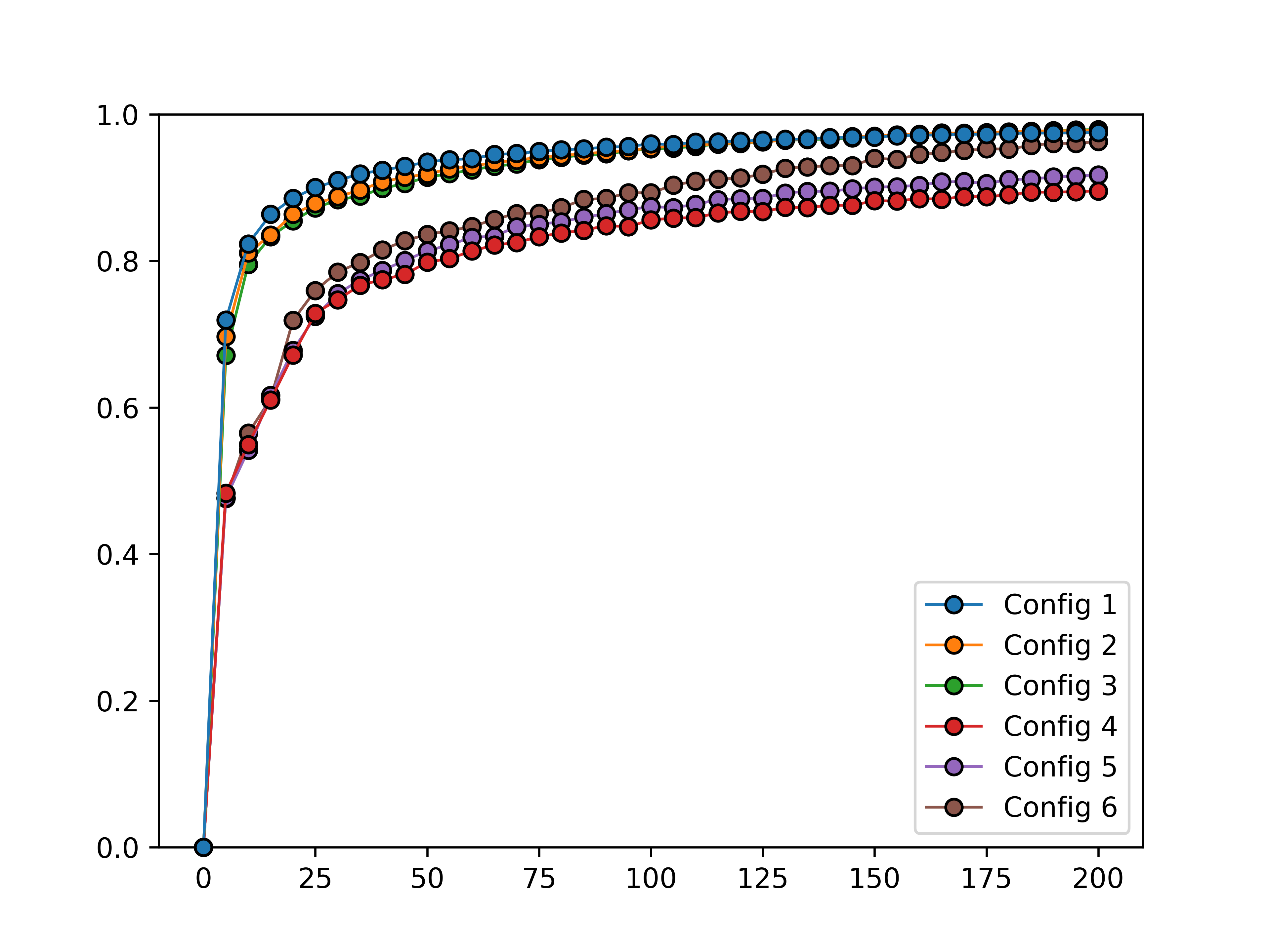} 
	\caption{Experimental results for the train dataset. The horizontal axis represents the number of training rounds, and the vertical axis represents the accuracy of the experiment. Our base configuration (Config 1) includes global features, static information, and all input features. Subsequent configurations build upon the previous one: Config 2 removes global features, Config 3 removes static information, Config 4 reduces input channels to just the center point, Config 5 increases channels to 256, and Config 6 further increases channels to 512.} 
	\label{fig:abl:2} 
\end{figure}

The experimental results for the test dataset are shown in Figure~\ref{fig:abl:1}, and the experimental results for the training dataset are shown in Figure~\ref{fig:abl:2}.
The horizontal axis is the number of training rounds, and the vertical axis is the accuracy of the experiment.
From Figure~\ref{fig:abl:1} and Figure~\ref{fig:abl:2}, we can see that our benchmark model performs the best, followed by the network with dynamic global information and static information removed, which shows a drop of about 1\% each.

More noteworthy is that in our experiments with only point information as input, the network performance shows a drop of at least 10\%.
This may be because we do not optimize the network, or it may be due to the decrease in the number of channels in the network.
So we try to increase the channel count, and we can see in Figure~\ref{fig:abl:2} that as the channel count goes up, the network is getting better and better in the training set and even approaching our model, but the performance is not improving in the test set.

So, we can conclude that adding more input feature information can improve the training speed and reduce overfitting compared to the method of using only point information. However, due to the information of the area normal, it can not improve the final accuracy, which may be the limitation of the method; other information, such as texture, mapping, etc., could be more effective in improving the accuracy.
Secondly, adding static and global features to the network can also slightly reduce the overfitting phenomenon.
It should be noted that when adding information with much noise, the performance may be degraded, so the proper data is better than data with errors.

\subsection{Effect of the number of neighbors}

\begin{table}[t]\footnotesize
	\centering

	\caption{Ablation experiments on the number of neighbors.}
	\label{table:ab-neighborhoods}
	\begin{tabular}{cc}
		\toprule

		Method   & Face Accuracy(\%) \\ 

		\midrule
		1024, 10 & 72.73             \\
		1024, 20 & 91.74             \\
		1024, 40 & 92.36             \\

		2048, 10 & 89.87             \\
		2048, 20 & 90.60             \\
		2048, 40 & 92.15             \\

		full, 10 & 92.95             \\
		full, 20 & \textbf{93.21}    \\
		full, 40 & \textbf{93.28}    \\
		\bottomrule
	\end{tabular}
\end{table}


The $k$ value of constructing the local graph structure has a significant influence on the extracted features. Similarly, the number of samples is also a factor to consider.
Therefore, it is essential to determine the influence of the $k$ value and the sampling number in this experiment.

We conduct nine sets of experiments to explore the impact of choosing different $k$ values on the part segmentation result in difference sampling num, which can be seen in Table~\ref{table:ab-neighborhoods}.
The result could be better if the sampling number is 1024 and the $k$ is 10. One of the reasons is that the neighborhoods of each center node are small and cannot fully interact with the neighbor node.
When the value of $k$ increases to 20, the result significantly improves, confirming our viewpoint.
When we increase the value of $k$ to 40, we notice a slight improvement in the results. However, the improvement could be more significant for two reasons. Firstly, it could be that the neighboring points have already reached the critical point. Secondly, increasing the number of neighboring points could introduce more noise, which could adversely affect the training of the model.

When using 2048 samples, we find that the change in $k$ value is less significant than when the samples are 1024.
Our method adds more meaningful information by increasing the number of samples, improving performance when the $k$ value is small. This is a way to eliminate the dependence on the $k$ value.

Some results with more samples under the same $k$ value may be because there are many more samples, but the relative density is lower if $k$ is kept the same. However, the overall results are similar, and our method adds more meaningful information when the $k$ value increases.

In light of the specific requirements of our experiments, we choose this scheme.
However, it is essential to note that the most appropriate approach may vary depending on the specific circumstances and demands of the situation.

\subsection{Selection of aggregation function}

\begin{table}[t]\footnotesize
	\centering

	\caption{Selection of different aggregation function.}
	\label{table:ab-aggr}
	\begin{tabular}{cc}
		\toprule

		Method & Face Accuracy(\%) \\ 

		\midrule
		max    & \textbf{93.2}     \\
		mean   & 92.0              \\
		sum    & 35.8              \\
		\bottomrule
	\end{tabular}
\end{table}





Due to the unordered nature of 3D data, it is crucial to employ symmetric functions like max, sum, or mean for feature computation to prevent significant issues arising from subsequent transformations. This approach is widely adopted in previous works~\cite{PointNetpp,charles_pointnet_2017}.
In our work, we also utilize symmetric functions to aggregate edge features and update the features of each central node.
We experiment with various aggregation functions, including max, sum, and mean.
The max function identifies the most significant feature among nodes within the local neighborhood.
The sum function combines the features of all nodes in the neighborhood.
The mean function normalizes the sum of features by dividing it by the number of nodes in the neighborhood.

Table~\ref{table:ab-aggr} shows the results of our selection of different aggregation functions on a part segmentation task in a human body dataset with 1024 nodes and 20 neighborhoods.
The sum aggregation method consistently fails to achieve satisfactory training performance, suggesting potential underlying issues within the network architecture.
While the max and mean pooling methods exhibit superior performance, the maximum pooling function, despite its theoretical loss of non-dominant feature information, effectively preserves the most salient features, as demonstrated by empirical evidence. Consequently, we opt for the max pooling function as the most effective edge feature aggregation technique.


\subsection{Impact on the number of hidden dimension}

\begin{table}[t]\footnotesize
	\centering

	\caption{Ablation experiments on the number of hidden dimension.}
	\begin{tabular}{cc}
		\toprule

		Method & Face Accuracy(\%) \\

		\midrule
		16     & 85.5              \\
		32     & 89.5              \\
		64     & 91.6              \\
		96     & 92.4              \\
		128    & \textbf{93.2}     \\
		192    & \textbf{93.2}     \\
		256    & 92.9              \\
		\bottomrule
	\end{tabular}
\end{table}

In the case of 16 hidden layers, the results are the worst, but not so bad that they cannot be trained.
This may be because the amount of data is small, or it may be because 16 layers, although they seem relatively small, are insufficient in this case.
In the case of 32 hidden layers, the result improves a lot, about 5\%, which shows that 16 hidden layers are still insufficient for this data set.
When the number of hidden layers is increased to 64, the accuracy improves by about 2\%, which is an improvement, but the trend is downward.
If the number of layers is doubled to 128, the accuracy also improves by 2\%, but only up to 192 layers.
However, when the number of layers is doubled to 192, there is no improvement in accuracy, and it reaches its limit.
After that, the accuracy rate decreases slightly when the hidden layer is increased to 256 layers. One reason may be that the network parameters must be more extensive and accessible for training. On the other hand, it may be due to the network parameters being too considerable and overfitting file aggravation.

\subsection{Discussion}

Our experiments show that adding simple computational input features does not necessarily improve performance.
This is because adding more data can also introduce more noise and complexity, making it more difficult for the model to learn the underlying patterns in the data and make accurate predictions.
Adding more data can sometimes lead to decreased performance; the additional data may have more noise than the original.
In the worst case, noise can lead to model failure. The model may learn to predict the noise instead of the signal.
However, the addition of dynamic global features steadily improves performance.
Dynamic global features are calculated from all features in a shape and aim to represent the shape.
Adding dynamic global features to a dataset can help the model learn more complex patterns in the data, leading to improved performance on tasks like segmentation.
Overall, adding more data only sometimes results in improving performance. It may even introduce additional noise and reduce the performance. However, adding dynamic global features can steadily improve the performance of machine learning models.

\section{Performance}
\label{sec:performance}

The algorithms in this paper are implemented using python, pytorch, and pytorch geometry~\cite{pyg}.
Our method's performance is measured on a PC with  Intel Core i7-12700K, 32 GB RAM, and NVIDIA GeForce 3090 GPU.
The running time of our method is shown in Table~\ref{table:runtime}. It shows that the running time of our method is mainly distributed in the training phase of the network, and the evaluation is speedy.


\begin{table}[t]\footnotesize
	\centering
	\caption{The running time (minutes) on our method.}
	\label{table:runtime}
	\begin{tabularx}{\columnwidth}{cccc}
		\toprule
		                         & \parbox{0.22\columnwidth}{Pre-Process\centering} & \parbox{0.22\columnwidth}{Network\centering         \\Training\centering} & \parbox{0.22\columnwidth}{Network\centering\\Evaluation\centering\\} \\

		\midrule
		total time & 82                                             & 218                                       & 0.5   \\
		average time & 0.2                                            & 0.54                                      & 0.001 \\
		\bottomrule
	\end{tabularx}
\end{table}


\section{Conclusion}
\label{sec:conclusion}

Even though the proposed network employs various data enhancement techniques to improve generalization, the network output can still be significantly biased for incorrectly modeled data, such as reversed face direction inputs. This type of data is usually not routinely considered, but some datasets suffer from such problems, leading to a decrease in network accuracy. Future research will focus on improving the robustness of the network to input perturbations, and one possible option is to incorporate error data into the data enhancement strategy.
In addition, our method has no constraints on the geometric or labeling characteristics, which leads to some small error regions like islands in the segmentation, and we can consider adding relevant constraints to minimize this situation to increase the accuracy.



In this paper, we propose a novel framework called InfoGNN for fast and scalable learning of end-to-end 3D mesh representations.
Our method treats 3D mesh models as graphs and utilizes graph neural networks to efficiently process irregular 3D mesh data. In addition, InfoGNN fully incorporates a variety of inputs, where the inclusion of dynamic global features notably improves the model performance.
Evaluation results on publicly available datasets show that InfoGNN achieves competitive performance and demonstrates the potential for a broader range of applications.

\section*{Acknowledgments}\label{sec:acknowledgements}
This work is supported by the National Natural Science Foundation of China (62172356, 61872321), Zhejiang Provincial Natural Science Foundation of China (LY22F020026),
the Ningbo Major Special Projects of the ``Science and Technology Innovation 2025'' (2020Z005, 2020Z007, 2021Z012).

\bibliography{bibTex/bibfile}

\providecommand{\latin}[1]{#1}
\makeatletter
\providecommand{\doi}
  {\begingroup\let\do\@makeother\dospecials
  \catcode`\{=1 \catcode`\}=2 \doi@aux}
\providecommand{\doi@aux}[1]{\endgroup\texttt{#1}}
\makeatother
\providecommand*\mcitethebibliography{\thebibliography}
\csname @ifundefined\endcsname{endmcitethebibliography}
  {\let\endmcitethebibliography\endthebibliography}{}
\begin{mcitethebibliography}{40}
\providecommand*\natexlab[1]{#1}
\providecommand*\mciteSetBstSublistMode[1]{}
\providecommand*\mciteSetBstMaxWidthForm[2]{}
\providecommand*\mciteBstWouldAddEndPuncttrue
  {\def\EndOfBibitem{\unskip.}}
\providecommand*\mciteBstWouldAddEndPunctfalse
  {\let\EndOfBibitem\relax}
\providecommand*\mciteSetBstMidEndSepPunct[3]{}
\providecommand*\mciteSetBstSublistLabelBeginEnd[3]{}
\providecommand*\EndOfBibitem{}
\mciteSetBstSublistMode{f}
\mciteSetBstMaxWidthForm{subitem}{(\alph{mcitesubitemcount})}
\mciteSetBstSublistLabelBeginEnd
  {\mcitemaxwidthsubitemform\space}
  {\relax}
  {\relax}

\bibitem[Yoon and Kim(2012)Yoon, and Kim]{Yoon_Kim_2012}
Yoon,~D.; Kim,~K.-J. {3D} game model and texture generation using interactive
  genetic algorithm. Proceedings of the Workshop at SIGGRAPH Asia. 2012; p
  53–58\relax
\mciteBstWouldAddEndPuncttrue
\mciteSetBstMidEndSepPunct{\mcitedefaultmidpunct}
{\mcitedefaultendpunct}{\mcitedefaultseppunct}\relax
\EndOfBibitem
\bibitem[Błaszczyk \latin{et~al.}(2021)Błaszczyk, Jabbar, Szmyd, and
  Radek]{Błaszczyk_Jabbar_Szmyd_Radek_2021}
Błaszczyk,~M.; Jabbar,~R.; Szmyd,~B.; Radek,~M. {3D} Printing of Rapid,
  Low-Cost and Patient-Specific Models of Brain Vasculature for Use in
  Preoperative Planning in Clipping of Intracranial Aneurysms. \emph{Journal of
  Clinical Medicine} \textbf{2021}, \emph{10}, 1201\relax
\mciteBstWouldAddEndPuncttrue
\mciteSetBstMidEndSepPunct{\mcitedefaultmidpunct}
{\mcitedefaultendpunct}{\mcitedefaultseppunct}\relax
\EndOfBibitem
\bibitem[Nozawa \latin{et~al.}(2019)Nozawa, Shum, Ho, and
  Morishima]{Nozawa_Shum_Ho_Morishima_2019}
Nozawa,~N.; Shum,~H. P.~H.; Ho,~E. S.~L.; Morishima,~S. {3D} Car Shape
  Reconstruction from a Single Sketch Image. Proceedings of the 12th ACM
  SIGGRAPH Conference on Motion, Interaction and Games. 2019; p~2\relax
\mciteBstWouldAddEndPuncttrue
\mciteSetBstMidEndSepPunct{\mcitedefaultmidpunct}
{\mcitedefaultendpunct}{\mcitedefaultseppunct}\relax
\EndOfBibitem
\bibitem[Lin \latin{et~al.}(2023)Lin, Gao, Tang, Takikawa, Zeng, Huang, Kreis,
  Fidler, Liu, and Lin]{lin2023magic3d}
Lin,~C.-H.; Gao,~J.; Tang,~L.; Takikawa,~T.; Zeng,~X.; Huang,~X.; Kreis,~K.;
  Fidler,~S.; Liu,~M.-Y.; Lin,~T.-Y. {Magic3D}: High-Resolution Text-to-{3D}
  Content Creation. Proceedings of the IEEE/CVF Conference on Computer Vision
  and Pattern Recognition. 2023; pp 300--309\relax
\mciteBstWouldAddEndPuncttrue
\mciteSetBstMidEndSepPunct{\mcitedefaultmidpunct}
{\mcitedefaultendpunct}{\mcitedefaultseppunct}\relax
\EndOfBibitem
\bibitem[Gal and Cohen-Or(2006)Gal, and Cohen-Or]{gal2006salient}
Gal,~R.; Cohen-Or,~D. Salient geometric features for partial shape matching and
  similarity. \emph{ACM Transactions on Graphics} \textbf{2006}, \emph{25},
  130--150\relax
\mciteBstWouldAddEndPuncttrue
\mciteSetBstMidEndSepPunct{\mcitedefaultmidpunct}
{\mcitedefaultendpunct}{\mcitedefaultseppunct}\relax
\EndOfBibitem
\bibitem[Shapira \latin{et~al.}(2010)Shapira, Shalom, Shamir, Cohen-Or, and
  Zhang]{shapira2010contextual}
Shapira,~L.; Shalom,~S.; Shamir,~A.; Cohen-Or,~D.; Zhang,~H. Contextual part
  analogies in {3D} objects. \emph{International Journal of Computer Vision}
  \textbf{2010}, \emph{89}, 309--326\relax
\mciteBstWouldAddEndPuncttrue
\mciteSetBstMidEndSepPunct{\mcitedefaultmidpunct}
{\mcitedefaultendpunct}{\mcitedefaultseppunct}\relax
\EndOfBibitem
\bibitem[Guo \latin{et~al.}(2015)Guo, Zou, and Chen]{guo20153d}
Guo,~K.; Zou,~D.; Chen,~X. {3D} mesh labeling via deep convolutional neural
  networks. \emph{ACM Transactions on Graphics} \textbf{2015}, \emph{35},
  1--12\relax
\mciteBstWouldAddEndPuncttrue
\mciteSetBstMidEndSepPunct{\mcitedefaultmidpunct}
{\mcitedefaultendpunct}{\mcitedefaultseppunct}\relax
\EndOfBibitem
\bibitem[Sarkar \latin{et~al.}(2018)Sarkar, Hampiholi, Varanasi, and
  Stricker]{Sarkar_Hampiholi_Varanasi_Stricker_2018}
Sarkar,~K.; Hampiholi,~B.; Varanasi,~K.; Stricker,~D. Learning {3D} Shapes as
  Multi-Layered Height-maps using {2D} Convolutional Networks. Proceedings of
  the European Conference on Computer Vision. 2018; p 74–89\relax
\mciteBstWouldAddEndPuncttrue
\mciteSetBstMidEndSepPunct{\mcitedefaultmidpunct}
{\mcitedefaultendpunct}{\mcitedefaultseppunct}\relax
\EndOfBibitem
\bibitem[Hanocka \latin{et~al.}(2019)Hanocka, Hertz, Fish, Giryes, Fleishman,
  and Cohen-Or]{hanocka_Meshcnn_2019}
Hanocka,~R.; Hertz,~A.; Fish,~N.; Giryes,~R.; Fleishman,~S.; Cohen-Or,~D.
  {MeshCNN}: A network with an edge. \emph{ACM Transactions on Graphics}
  \textbf{2019}, \emph{38}, 1--12\relax
\mciteBstWouldAddEndPuncttrue
\mciteSetBstMidEndSepPunct{\mcitedefaultmidpunct}
{\mcitedefaultendpunct}{\mcitedefaultseppunct}\relax
\EndOfBibitem
\bibitem[Wang \latin{et~al.}(2019)Wang, Sun, Liu, Sarma, Bronstein, and
  Solomon]{DGCNN}
Wang,~Y.; Sun,~Y.; Liu,~Z.; Sarma,~S.~E.; Bronstein,~M.~M.; Solomon,~J.~M.
  Dynamic {graph} {CNN} for {learning} on {point} {clouds}. \emph{ACM
  Transactions on Graphics} \textbf{2019}, \emph{38}, 1--12\relax
\mciteBstWouldAddEndPuncttrue
\mciteSetBstMidEndSepPunct{\mcitedefaultmidpunct}
{\mcitedefaultendpunct}{\mcitedefaultseppunct}\relax
\EndOfBibitem
\bibitem[Shapira \latin{et~al.}(2008)Shapira, Shamir, and
  Cohen-Or]{Shapira_Shamir_Cohen-Or_2008}
Shapira,~L.; Shamir,~A.; Cohen-Or,~D. Consistent mesh partitioning and
  skeletonisation using the shape diameter function. \emph{The Visual Computer}
  \textbf{2008}, \emph{24}, 249–259\relax
\mciteBstWouldAddEndPuncttrue
\mciteSetBstMidEndSepPunct{\mcitedefaultmidpunct}
{\mcitedefaultendpunct}{\mcitedefaultseppunct}\relax
\EndOfBibitem
\bibitem[Bronstein and Kokkinos(2010)Bronstein, and
  Kokkinos]{Bronstein_Kokkinos_2010}
Bronstein,~M.~M.; Kokkinos,~I. Scale-invariant heat kernel signatures for
  non-rigid shape recognition. 2010 IEEE Computer Society Conference on
  Computer Vision and Pattern Recognition. 2010; pp 1704--1711\relax
\mciteBstWouldAddEndPuncttrue
\mciteSetBstMidEndSepPunct{\mcitedefaultmidpunct}
{\mcitedefaultendpunct}{\mcitedefaultseppunct}\relax
\EndOfBibitem
\bibitem[Raviv \latin{et~al.}(2010)Raviv, Bronstein, Bronstein, and
  Kimmel]{Raviv_Bronstein_Bronstein_Kimmel_2010}
Raviv,~D.; Bronstein,~M.~M.; Bronstein,~A.~M.; Kimmel,~R. Volumetric heat
  kernel signatures. Proceedings of the ACM workshop on {3D} object retrieval.
  2010; p 39–44\relax
\mciteBstWouldAddEndPuncttrue
\mciteSetBstMidEndSepPunct{\mcitedefaultmidpunct}
{\mcitedefaultendpunct}{\mcitedefaultseppunct}\relax
\EndOfBibitem
\bibitem[Xie \latin{et~al.}(2014)Xie, Xu, Liu, and
  Xiong]{Xie_Xu_Liu_Xiong_2014}
Xie,~Z.; Xu,~K.; Liu,~L.; Xiong,~Y. {3D} Shape Segmentation and Labeling via
  Extreme Learning Machine. Proceedings of the Symposium on Geometry
  Processing. 2014; p 85–95\relax
\mciteBstWouldAddEndPuncttrue
\mciteSetBstMidEndSepPunct{\mcitedefaultmidpunct}
{\mcitedefaultendpunct}{\mcitedefaultseppunct}\relax
\EndOfBibitem
\bibitem[Su \latin{et~al.}(2015)Su, Maji, Kalogerakis, and
  Learned-Miller]{Su_Maji_Kalogerakis_Learned-Miller_2015}
Su,~H.; Maji,~S.; Kalogerakis,~E.; Learned-Miller,~E. Multi-view Convolutional
  Neural Networks for {3D} Shape Recognition. Proceedings of the IEEE
  International Conference on Computer Vision. 2015; pp 945--953\relax
\mciteBstWouldAddEndPuncttrue
\mciteSetBstMidEndSepPunct{\mcitedefaultmidpunct}
{\mcitedefaultendpunct}{\mcitedefaultseppunct}\relax
\EndOfBibitem
\bibitem[Kalogerakis \latin{et~al.}(2017)Kalogerakis, Averkiou, Maji, and
  Chaudhuri]{Kalogerakis_Averkiou_Maji_Chaudhuri_2016}
Kalogerakis,~E.; Averkiou,~M.; Maji,~S.; Chaudhuri,~S. {3D} Shape Segmentation
  with Projective Convolutional Networks. IEEE Conference on Computer Vision
  and Pattern Recognition. 2017; pp 6630--6639\relax
\mciteBstWouldAddEndPuncttrue
\mciteSetBstMidEndSepPunct{\mcitedefaultmidpunct}
{\mcitedefaultendpunct}{\mcitedefaultseppunct}\relax
\EndOfBibitem
\bibitem[Shen \latin{et~al.}(2018)Shen, Feng, Yang, and Tian]{KCnet}
Shen,~Y.; Feng,~C.; Yang,~Y.; Tian,~D. Mining Point Cloud Local Structures by
  Kernel Correlation and Graph Pooling. Proceedings of the IEEE Conference on
  Computer Vision and Pattern Recognition. 2018; pp 4548--4557\relax
\mciteBstWouldAddEndPuncttrue
\mciteSetBstMidEndSepPunct{\mcitedefaultmidpunct}
{\mcitedefaultendpunct}{\mcitedefaultseppunct}\relax
\EndOfBibitem
\bibitem[Lahav and Tal(2020)Lahav, and Tal]{lahav_Meshwalker_2020}
Lahav,~A.; Tal,~A. {MeshWalker}: Deep mesh understanding by random walks.
  \emph{ACM Transactions on Graphics} \textbf{2020}, \emph{39}, 1--13\relax
\mciteBstWouldAddEndPuncttrue
\mciteSetBstMidEndSepPunct{\mcitedefaultmidpunct}
{\mcitedefaultendpunct}{\mcitedefaultseppunct}\relax
\EndOfBibitem
\bibitem[Li \latin{et~al.}(2023)Li, He, Fan, and Song]{TPNet}
Li,~P.; He,~F.; Fan,~B.; Song,~Y. {TPNet}: A novel mesh analysis method via
  topology preservation and perception enhancement. \emph{Computer Aided
  Geometric Design} \textbf{2023}, \emph{104}, 11\relax
\mciteBstWouldAddEndPuncttrue
\mciteSetBstMidEndSepPunct{\mcitedefaultmidpunct}
{\mcitedefaultendpunct}{\mcitedefaultseppunct}\relax
\EndOfBibitem
\bibitem[Charles \latin{et~al.}(2017)Charles, Su, Kaichun, and
  Guibas]{charles_pointnet_2017}
Charles,~R.~Q.; Su,~H.; Kaichun,~M.; Guibas,~L.~J. {PointNet}: {Deep}
  {learning} on {point} {sets} for {3D} {classification} and {segmentation}.
  Proceedings of the IEEE Conference on Computer Vision and Pattern
  Recognition. 2017; pp 652--660\relax
\mciteBstWouldAddEndPuncttrue
\mciteSetBstMidEndSepPunct{\mcitedefaultmidpunct}
{\mcitedefaultendpunct}{\mcitedefaultseppunct}\relax
\EndOfBibitem
\bibitem[Anguelov \latin{et~al.}(2005)Anguelov, Srinivasan, Koller, Thrun,
  Rodgers, and Davis]{SCAPE}
Anguelov,~D.; Srinivasan,~P.; Koller,~D.; Thrun,~S.; Rodgers,~J.; Davis,~J.
  {SCAPE}: Shape completion and animation of people. \emph{ACM Transactions on
  Graphics} \textbf{2005}, \emph{24}, 408--416\relax
\mciteBstWouldAddEndPuncttrue
\mciteSetBstMidEndSepPunct{\mcitedefaultmidpunct}
{\mcitedefaultendpunct}{\mcitedefaultseppunct}\relax
\EndOfBibitem
\bibitem[Bogo \latin{et~al.}(2014)Bogo, Romero, Loper, and Black]{FAUST}
Bogo,~F.; Romero,~J.; Loper,~M.; Black,~M.~J. {FAUST}: Dataset and evaluation
  for {3D} mesh registration. Proceedings IEEE Conference on Computer Vision
  and Pattern Recognition. 2014; pp 3794 --3801\relax
\mciteBstWouldAddEndPuncttrue
\mciteSetBstMidEndSepPunct{\mcitedefaultmidpunct}
{\mcitedefaultendpunct}{\mcitedefaultseppunct}\relax
\EndOfBibitem
\bibitem[Vlasic \latin{et~al.}(2008)Vlasic, Baran, Matusik, and
  Popovi{\'c}]{Vlasic2008ArticulatedMA}
Vlasic,~D.; Baran,~I.; Matusik,~W.; Popovi{\'c},~J. Articulated mesh animation
  from multi-view silhouettes. \emph{ACM Transactions on Graphics}
  \textbf{2008}, \emph{27}, 1–9\relax
\mciteBstWouldAddEndPuncttrue
\mciteSetBstMidEndSepPunct{\mcitedefaultmidpunct}
{\mcitedefaultendpunct}{\mcitedefaultseppunct}\relax
\EndOfBibitem
\bibitem[Giorgi \latin{et~al.}(2008)Giorgi, Biasotti, and Paraboschi]{SHREC07}
Giorgi,~D.; Biasotti,~S.; Paraboschi,~L. Shape Retrieval Contest 2007:
  Watertight Models Track. \emph{SHREC Competition} \textbf{2008}, \emph{8},
  7\relax
\mciteBstWouldAddEndPuncttrue
\mciteSetBstMidEndSepPunct{\mcitedefaultmidpunct}
{\mcitedefaultendpunct}{\mcitedefaultseppunct}\relax
\EndOfBibitem
\bibitem[Kalogerakis \latin{et~al.}(2010)Kalogerakis, Hertzmann, and
  Singh]{Kalogerakis2010}
Kalogerakis,~E.; Hertzmann,~A.; Singh,~K. Learning {3D} Mesh Segmentation and
  Labeling. \emph{ACM Transactions on Graphics} \textbf{2010}, \emph{29},
  12\relax
\mciteBstWouldAddEndPuncttrue
\mciteSetBstMidEndSepPunct{\mcitedefaultmidpunct}
{\mcitedefaultendpunct}{\mcitedefaultseppunct}\relax
\EndOfBibitem
\bibitem[Haim \latin{et~al.}(2019)Haim, Segol, Ben-Hamu, Maron, and
  Lipman]{SNGC}
Haim,~N.; Segol,~N.; Ben-Hamu,~H.; Maron,~H.; Lipman,~Y. Surface Networks via
  General Covers. Proceedings of the IEEE/CVF International Conference on
  Computer Vision. 2019; pp 632--641\relax
\mciteBstWouldAddEndPuncttrue
\mciteSetBstMidEndSepPunct{\mcitedefaultmidpunct}
{\mcitedefaultendpunct}{\mcitedefaultseppunct}\relax
\EndOfBibitem
\bibitem[Maron \latin{et~al.}(2017)Maron, Galun, Aigerman, Trope, Dym, Yumer,
  Kim, and Lipman]{ToricCover}
Maron,~H.; Galun,~M.; Aigerman,~N.; Trope,~M.; Dym,~N.; Yumer,~E.; Kim,~V.~G.;
  Lipman,~Y. Convolutional Neural Networks on Surfaces via Seamless Toric
  Covers. \emph{ACM Transactions on Graphics} \textbf{2017}, \emph{36},
  10\relax
\mciteBstWouldAddEndPuncttrue
\mciteSetBstMidEndSepPunct{\mcitedefaultmidpunct}
{\mcitedefaultendpunct}{\mcitedefaultseppunct}\relax
\EndOfBibitem
\bibitem[Masci \latin{et~al.}(2015)Masci, Boscaini, Bronstein, and
  Vandergheynst]{GCNN}
Masci,~J.; Boscaini,~D.; Bronstein,~M.~M.; Vandergheynst,~P. Geodesic
  Convolutional Neural Networks on Riemannian Manifolds. Proceedings of the
  IEEE International Conference on Computer Vision Workshops. 2015; pp
  832--840\relax
\mciteBstWouldAddEndPuncttrue
\mciteSetBstMidEndSepPunct{\mcitedefaultmidpunct}
{\mcitedefaultendpunct}{\mcitedefaultseppunct}\relax
\EndOfBibitem
\bibitem[Poulenard and Ovsjanikov(2018)Poulenard, and Ovsjanikov]{MDGCNN}
Poulenard,~A.; Ovsjanikov,~M. Multi-directional Geodesic Neural Networks via
  Equivariant Convolution. 2018; pp 1--14\relax
\mciteBstWouldAddEndPuncttrue
\mciteSetBstMidEndSepPunct{\mcitedefaultmidpunct}
{\mcitedefaultendpunct}{\mcitedefaultseppunct}\relax
\EndOfBibitem
\bibitem[Qi \latin{et~al.}(2017)Qi, Yi, Su, and Guibas]{PointNetpp}
Qi,~C.~R.; Yi,~L.; Su,~H.; Guibas,~L.~J. {PointNet++}: Deep hierarchical
  feature learning on point sets in a metric space. Proceedings of the 31st
  International Conference on Neural Information Processing Systems. 2017; p
  5105–5114\relax
\mciteBstWouldAddEndPuncttrue
\mciteSetBstMidEndSepPunct{\mcitedefaultmidpunct}
{\mcitedefaultendpunct}{\mcitedefaultseppunct}\relax
\EndOfBibitem
\bibitem[Feng \latin{et~al.}(2019)Feng, Feng, You, Zhao, and Gao]{MeshNet}
Feng,~Y.; Feng,~Y.; You,~H.; Zhao,~X.; Gao,~Y. {MeshNet}: Mesh Neural Network
  for {3D} Shape Representation. Proceedings of the AAAI Conference on
  Artificial Intelligence. 2019; pp 8279--8286\relax
\mciteBstWouldAddEndPuncttrue
\mciteSetBstMidEndSepPunct{\mcitedefaultmidpunct}
{\mcitedefaultendpunct}{\mcitedefaultseppunct}\relax
\EndOfBibitem
\bibitem[Paul \latin{et~al.}(2023)Paul, Patterson, and Bouguila]{10229841}
Paul,~S.; Patterson,~Z.; Bouguila,~N. CrossMoCo: Multi-Modal Momentum
  Contrastive Learning for Point Cloud. 2023 20th Conference on Robots and
  Vision. 2023; pp 273--280\relax
\mciteBstWouldAddEndPuncttrue
\mciteSetBstMidEndSepPunct{\mcitedefaultmidpunct}
{\mcitedefaultendpunct}{\mcitedefaultseppunct}\relax
\EndOfBibitem
\bibitem[Klokov and Lempitsky(2017)Klokov, and Lempitsky]{KdNet}
Klokov,~R.; Lempitsky,~V. Escape from Cells: Deep Kd-networks for the
  recognition of {3D} point cloud models. IEEE International Conference on
  Computer Vision. 2017; pp 863--872\relax
\mciteBstWouldAddEndPuncttrue
\mciteSetBstMidEndSepPunct{\mcitedefaultmidpunct}
{\mcitedefaultendpunct}{\mcitedefaultseppunct}\relax
\EndOfBibitem
\bibitem[Atzmon \latin{et~al.}(2018)Atzmon, Maron, and Lipman]{PCNN}
Atzmon,~M.; Maron,~H.; Lipman,~Y. Point Convolutional Neural Networks by
  Extension Operators. \emph{ACM Transactions on Graphics} \textbf{2018},
  \emph{37}, 12\relax
\mciteBstWouldAddEndPuncttrue
\mciteSetBstMidEndSepPunct{\mcitedefaultmidpunct}
{\mcitedefaultendpunct}{\mcitedefaultseppunct}\relax
\EndOfBibitem
\bibitem[Lin \latin{et~al.}(2020)Lin, Huang, and Wang]{Lin_2020_CVPR}
Lin,~Z.-H.; Huang,~S.-Y.; Wang,~Y.-C.~F. Convolution in the Cloud: Learning
  Deformable Kernels in {3D} Graph Convolution Networks for Point Cloud
  Analysis. 2020 IEEE/CVF Conference on Computer Vision and Pattern
  Recognition. 2020; pp 1797--1806\relax
\mciteBstWouldAddEndPuncttrue
\mciteSetBstMidEndSepPunct{\mcitedefaultmidpunct}
{\mcitedefaultendpunct}{\mcitedefaultseppunct}\relax
\EndOfBibitem
\bibitem[Cheng \latin{et~al.}(2021)Cheng, Chen, He, Liu, and
  Bai]{cheng_pra-net_2021}
Cheng,~S.; Chen,~X.; He,~X.; Liu,~Z.; Bai,~X. {PRA}-{Net}: {Point}
  {relation}-{aware} {network} for {3D} {point} {cloud} {analysis}. \emph{IEEE
  Transactions on Image Processing} \textbf{2021}, \emph{30}, 4436--4448\relax
\mciteBstWouldAddEndPuncttrue
\mciteSetBstMidEndSepPunct{\mcitedefaultmidpunct}
{\mcitedefaultendpunct}{\mcitedefaultseppunct}\relax
\EndOfBibitem
\bibitem[Yi \latin{et~al.}(2016)Yi, Kim, Ceylan, Shen, Yan, Su, Lu, Huang,
  Sheffer, and Guibas]{shapenet2015}
Yi,~L.; Kim,~V.~G.; Ceylan,~D.; Shen,~I.-C.; Yan,~M.; Su,~H.; Lu,~C.;
  Huang,~Q.; Sheffer,~A.; Guibas,~L. A scalable active framework for region
  annotation in {3D} shape collections. \emph{ACM Transactions on Graphics}
  \textbf{2016}, \emph{35}, 12\relax
\mciteBstWouldAddEndPuncttrue
\mciteSetBstMidEndSepPunct{\mcitedefaultmidpunct}
{\mcitedefaultendpunct}{\mcitedefaultseppunct}\relax
\EndOfBibitem
\bibitem[Wu \latin{et~al.}(2015)Wu, Song, Khosla, Yu, Zhang, Tang, and
  Xiao]{modelnet}
Wu,~Z.; Song,~S.; Khosla,~A.; Yu,~F.; Zhang,~L.; Tang,~X.; Xiao,~J. {3D}
  ShapeNets: A deep representation for volumetric shapes. IEEE Conference on
  Computer Vision and Pattern Recognition. 2015; pp 1912--1920\relax
\mciteBstWouldAddEndPuncttrue
\mciteSetBstMidEndSepPunct{\mcitedefaultmidpunct}
{\mcitedefaultendpunct}{\mcitedefaultseppunct}\relax
\EndOfBibitem
\bibitem[Fey and Lenssen(2019)Fey, and Lenssen]{pyg}
Fey,~M.; Lenssen,~J.~E. Fast Graph Representation Learning with {PyTorch}
  Geometric. 2019; \url{http://arxiv.org/abs/1903.02428}\relax
\mciteBstWouldAddEndPuncttrue
\mciteSetBstMidEndSepPunct{\mcitedefaultmidpunct}
{\mcitedefaultendpunct}{\mcitedefaultseppunct}\relax
\EndOfBibitem
\end{mcitethebibliography}

\end{document}